\documentclass[runningheads]{llncs}
\usepackage{graphicx}
\usepackage{makecell}

\usepackage{tikz}
\usepackage{comment}
\usepackage{amsmath,amssymb} 
\usepackage{color}
\usepackage{booktabs}
\usepackage{arydshln}
\usepackage{multirow}
\usepackage{hyperref}

\hypersetup{colorlinks=true,linkcolor=blue,filecolor=magenta,urlcolor=cyan,
            pdftitle={Overleaf Example},pdfpagemode=FullScreen}

\usepackage[accsupp]{axessibility}  

\usepackage{caption}
\begin{document}

\pagestyle{headings}
\mainmatter
\def\ECCVSubNumber{1225}  

\newcommand{\blue}[1]{\textbf{\textcolor{mblue}{#1}}}
\newcommand{\red}[1]{\textcolor{red}{#1}}
\newcommand{\bred}[1]{\textbf{\textcolor{red}{#1}}}
\newcommand{\green}[1]{\textcolor{mgreen}{#1}}
\newcommand{\yellow}[1]{\textcolor{yellow}{#1}}
\newcommand{\gray}[1]{\textcolor{mgray}{#1}}
\newcommand{\supplement}{supplementary materials}
\newcommand{\textbfg}[1]{\textbf{\textcolor{mgreen}{#1}}}

\title{Quality Assessment in the Time Dimension: \\ Inter-frame Distortion and Time-series Attention} 

\title{FAST-VQA: Efficient End-to-end Video Quality Assessment with Fragment Sampling} 

%
\titlerunning{ECCV-22 submission ID \ECCVSubNumber} 
\authorrunning{ECCV-22 submission ID \ECCVSubNumber} 
\author{Anonymous ECCV submission}
\institute{Paper ID \ECCVSubNumber}

%


\titlerunning{FAST-VQA}

\author{Haoning Wu\inst{1,2,3} \and
Chaofeng Chen\inst{1,2} \and
Jingwen Hou\inst{2} \and
Liang Liao\inst{1,2} \and
Annan Wang\inst{1,2} \and
Wenxiu Sun\inst{3} \and
Qiong Yan\inst{3} \and
Weisi Lin\inst{2}}
\authorrunning{H. Wu \textit{et al.}}
%
\institute{S-Lab, Nanyang Technological University \and
School of Computer Science and Engineering, Nanyang Technological University \and
Sensetime Research and Tetras AI\\
\email{haoning001@e.ntu.edu.sg}
}

\newcommand{\frag}[0]{\textbf{{\textit{fragments}}}}

\newcommand{\cfchen}[1]{\textcolor{red}{CF: #1}}
\newcommand{\cfcomment}[1]{\textcolor{blue}{\emph{CF Comment: #1}}}
\newcommand{\LL}[1]{\textcolor{red}{\emph{LL: #1}}}


\maketitle

\definecolor{mgray}{gray}{0.45}
\definecolor{mred}{RGB}{238, 34, 12}
\definecolor{mgreen}{RGB}{1, 127, 0}
\definecolor{mblue}{RGB}{0, 0, 180}

\begin{abstract}
Current deep video quality assessment (VQA) methods are usually with high computational costs when evaluating high-resolution videos. This cost hinders them from learning better video-quality-related representations via end-to-end training. Existing approaches usually consider naive sampling to reduce the computational cost, such as \textit{resizing} and \textit{cropping}. However, they obviously corrupt quality-related information in videos and are thus not optimal to learn good representations for VQA. Therefore, there is an eager need to design a new quality-retained sampling scheme for VQA. In this paper, we propose Grid Mini-patch Sampling (GMS), which allows consideration of local quality by sampling patches at their raw resolution and covers global quality with contextual relations via mini-patches sampled in uniform grids. These mini-patches are spliced and aligned temporally, named as \frag. We further build the Fragment Attention Network (FANet) specially designed to accommodate \frag~as inputs.
Consisting of \frag~and FANet, the proposed FrAgment Sample Transformer for VQA (\textbf{FAST-VQA}) enables efficient end-to-end deep VQA and learns effective video-quality-related representations. It improves state-of-the-art accuracy by around \textbf{$10\%$} while reducing \textbf{$99.5\%$} FLOPs on 1080P high-resolution videos. The newly learned video-quality-related representations can also be transferred into smaller VQA datasets and boost the performance on these scenarios. Extensive experiments show that FAST-VQA has 
good performance on inputs of various resolutions while retaining high efficiency. We publish our code at \url{https://github.com/timothyhtimothy/FAST-VQA}.


\end{abstract}

\section{Introduction}
\label{sec:1}

\begin{figure}[]
    \centering
    \includegraphics[width=0.93\linewidth]{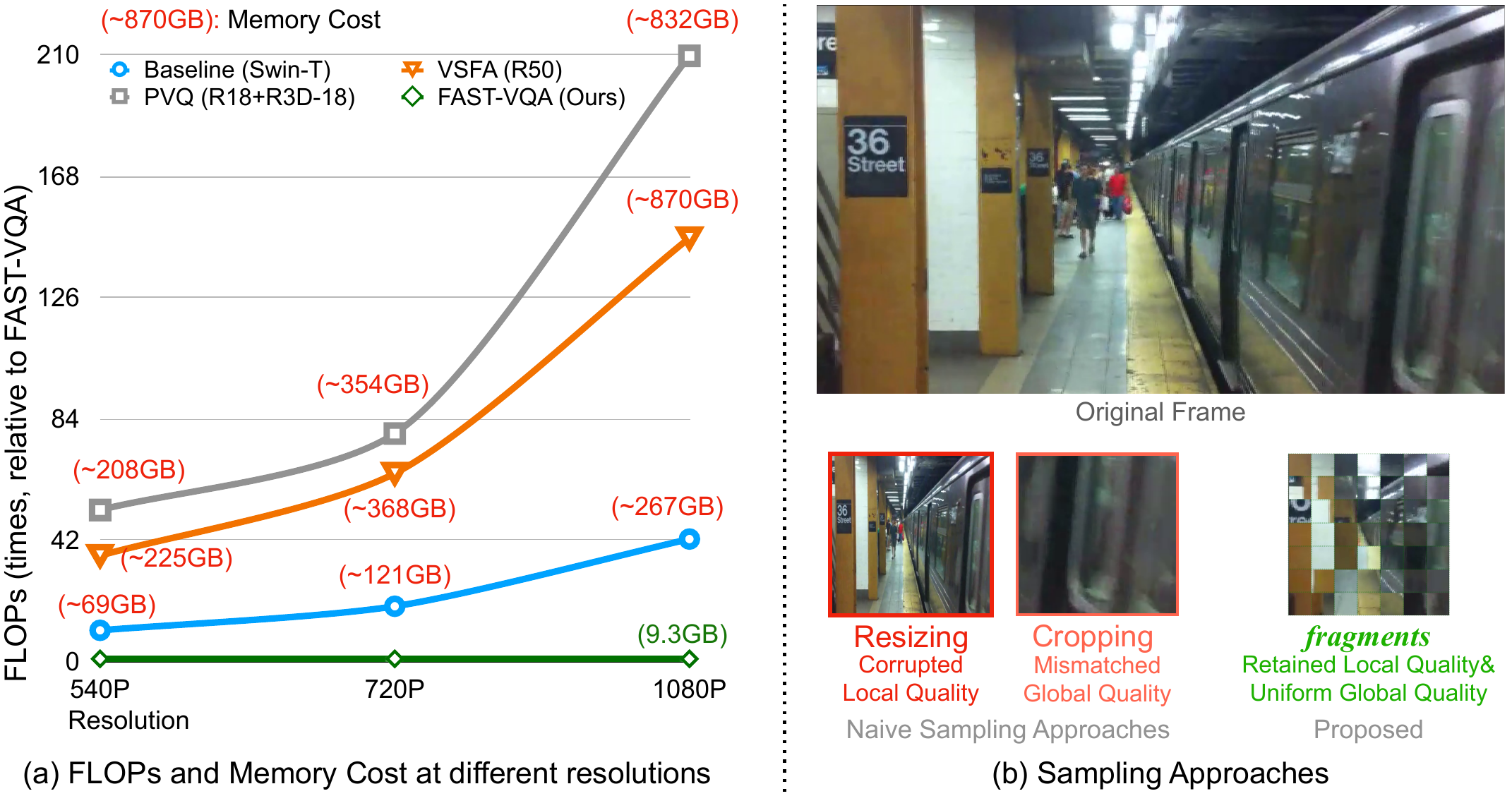}
    \vspace{-5pt}
    \caption{Motivation for \frag: (a) The computational cost (FLOPs\&Memory at Batch Size 4) for existing VQA methods is high especially on high-resolution videos. (b) Sampling approaches. Naive approaches such as \textit{resizing}~\cite{musiq,dbcnn} and \textit{cropping}~\cite{crop1,crop2} cannot preserve video quality well. Zoom in for clearer view.}
    \label{fig:1}
    \vspace{-19pt}
\end{figure}

More and more videos with a variety of contents are collected in-the-wild and uploaded to the Internet every day. With the growth of high-definition video recording devices, a growing proportion of these videos are in high resolution (e.g. $\geq 1080P$). Classical video quality assessment (VQA) algorithms based on handcrafted features are difficult to handle these videos with diverse content and degradation. In recent years, deep-learning-based VQA methods \cite{vsfa,mdtvsfa,pvq,mlsp,cnn+lstm,cnntlvqm} have shown better performance on in-the-wild VQA benchmarks \cite{vqc,kv1k,ytugc,pvq}. However, the computational cost of deep VQA methods increases quadratically when applied to high resolution videos, and a video of size $1080\times1920$ would require $\mathbf{42.5\times}$ floating point operations (FLOPs) than normal $224\times224$ inputs (as Fig.~\ref{fig:1}(a) shows), limiting these methods from practical applications. It is urgent to develop new VQA methods that are both effective and efficient.

Meanwhile, with high memory cost noted in Fig.~\ref{fig:1}(a), existing methods usually regress quality scores with \red{\textbf{fixed}} features extracted from pre-trained networks for classification tasks~\cite{he2016residual,irnv2,r3d} to alleviate memory shortage problem on GPUs instead of end-to-end training, preventing them from learning \textit{video-quality-related representations} that better represent quality information and limiting their accuracy. Existing approaches apply naive sampling on images or videos by resizing~\cite{musiq,dbcnn} or cropping~\cite{crop1,crop2} (as Fig.~\ref{fig:1}(b) shows) to reduce this cost and enable end-to-end training. However, they both cause artificial quality corruptions or changes during sampling, \emph{e.g.}, resizing corrupts local textures that are significant for predicting video quality, while cropping causes mismatched global quality with local regions. Moreover, the severity of these problems increases with the raw resolution of the video, making them unsuitable for VQA tasks.

\begin{figure}[htbp]
    \centering
    \includegraphics[width=\linewidth]{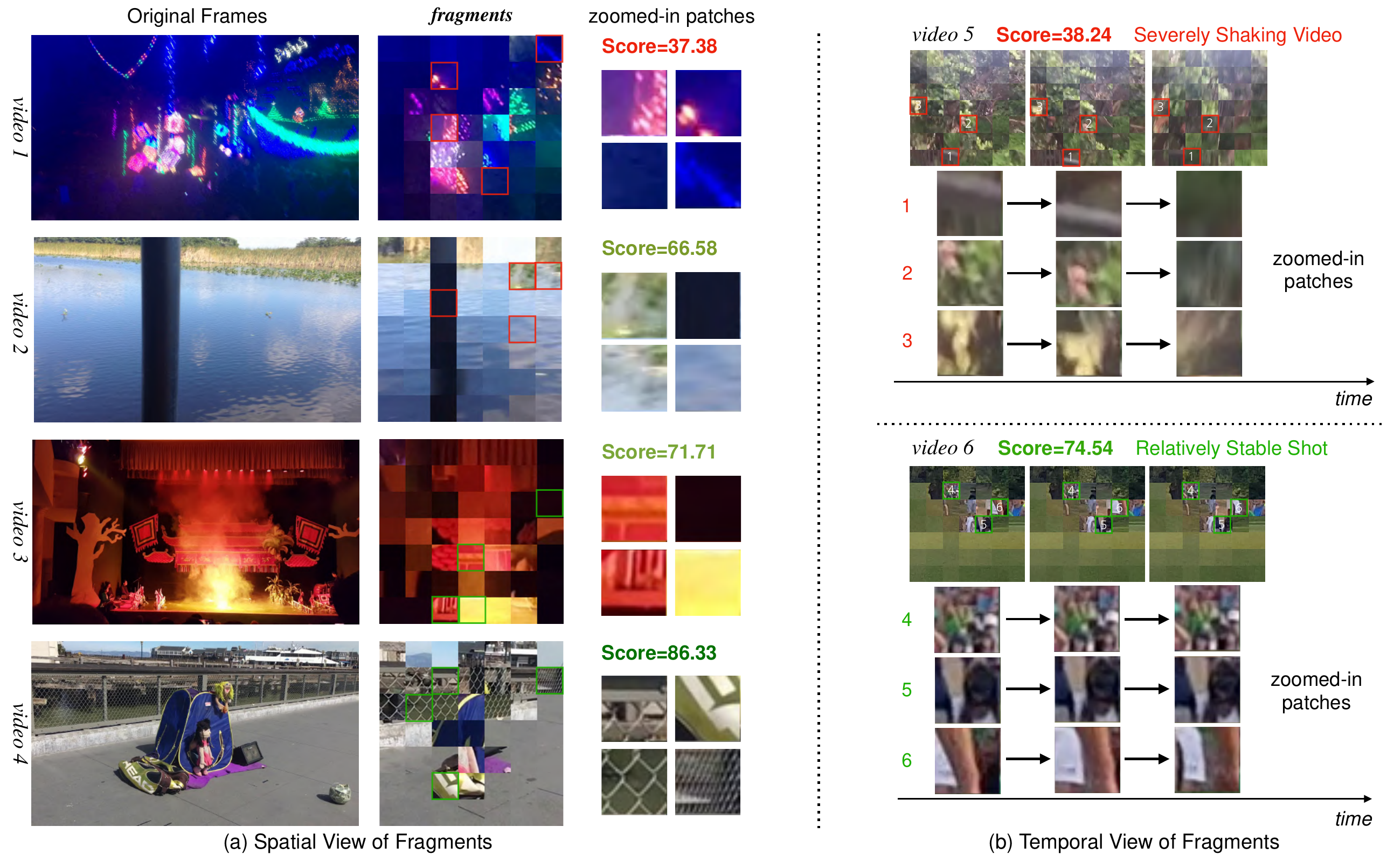}
    \caption{\textbf{\textit{Fragments}}, in spatial view (a) and temporal view (b). Zoom-in views of mini-patches show that \textbf{\textit{fragments}} can retain spatial local quality information (a), and spot temporal variations such as shaking across frames (b). In (a), spliced mini-patches also keep the global scene information of original frames.}
    \label{fig:2}
    \vspace{-20pt}
\end{figure}

To improve the practical efficiency and the training effectiveness of deep VQA methods, we propose a new sampling scheme, Grid Mini-patch Sampling (GMS), to retain the sensitivity to original video quality. GMS cuts videos into spatially uniform non-overlapping grids, randomly sample a mini-patch from each grid, and then splice mini-patches together. In temporal view, we constrain the position of mini-patches to align across frames, in order to ensure the sensitivity on temporal variations. We name these temporally aligned and spatially spliced mini-patches as \frag. As shown in Fig.~\ref{fig:2}, The proposed fragments can well preserve the sensitivity on both spatial and temporal quality. First, it preserves the local texture-related quality information (\emph{e.g.}, spot blurs happened in \textit{video 1/2}) by retaining the original resolution in patches. Second, benefiting from the globally uniformly sampled grids, it covers the global quality even though different regions have different qualities (\emph{e.g.}, \textit{video 3}). Third, by splicing the mini-patches, \frag~retains contextual relations of patches so that the model can learn global scene information of the original frames. At last, with temporal alignment, \frag~preserve temporal quality sensitivity by retaining the inter-frame variations in mini-patches from raw resolution, so they can be used to spot temporal distortions in videos and distinguish between severely shaking videos (\emph{e.g.}, \textit{video 5}) from relatively stable shots (\emph{e.g.}, \textit{video 6}).

However, it is non-trivial to build a network using the proposed \frag~as inputs. The network should follow two principles: 1) It should better extract the quality-related information preserved in \frag, including the retained local textures inside the raw resolution patches and the contextual relations between the spliced mini-patches; 2) It should distinguish the artificial discontinuity between mini-patches in \frag~from the authentic quality degradation in the original videos. Based on these two principles, we propose a Fragment Attention Network (FANet) with Video Swin Transformer Tiny (Swin-T)~\cite{swin3d}  as the backbone. Swin-T has a hierarchical structure and processes inputs with patch-wise operations, which is naturally suitable for proceeding with proposed \frag.

\begin{figure}[]
    \centering
    \includegraphics[width=0.92\linewidth]{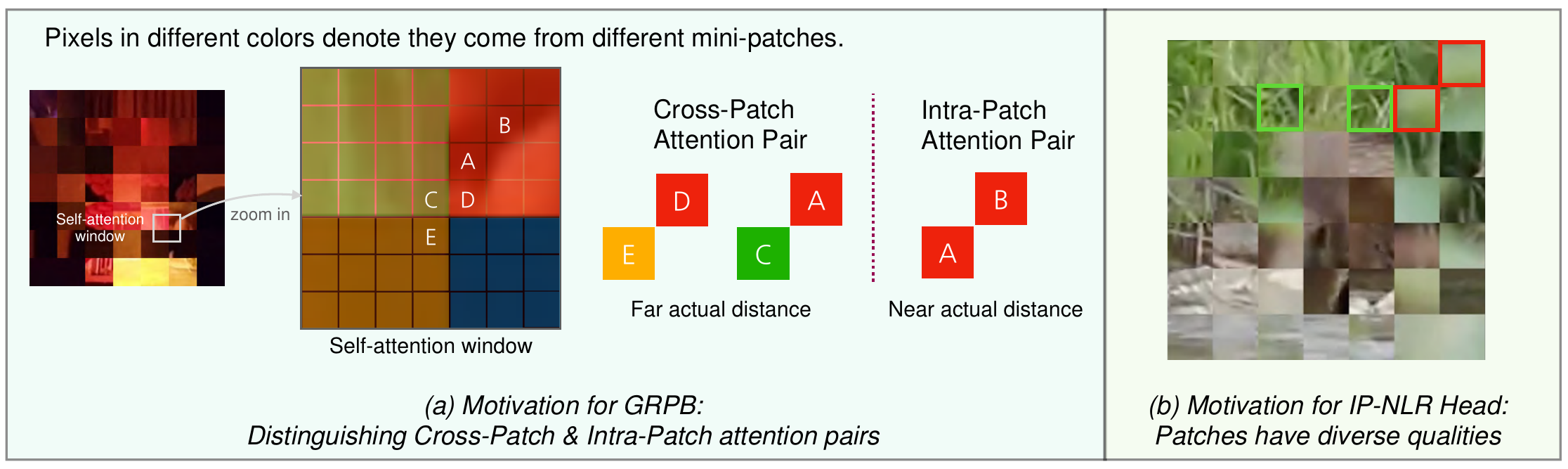}
    \vspace{-5pt}
    \caption{Motivation for the two proposed modules in FANet: (a) Gated Relative Position Biases (GRPB); (b) Intra-Patch Non-Linear Regression (IP-NLR) head. The structures for the two modules are illustrated in Fig.~\ref{fig:5}.}
    \label{fig:3}
    \vspace{-18pt}
\end{figure}

Furthermore, to avoid the negative impact of discontinuity between mini-patches on quality prediction, we propose two novel modules, \emph{i.e.}, Gated Relative Position Biases (GRPB) and Intra-Patch Non-Linear Regression (IP-NLR), to correct for the self-attention computation and the final score regression in the FANet respectively. Specifically, considering that some pairs in the same attention window might have the same relative position (\textit{e.g.}, Fig.~\ref{fig:3}(a) \red{A}-\green{C}, \red{D}-\yellow{E}, \red{A}-\red{B}), 
but the cross-patch attention pairs (\red{A}-\green{C}, \red{D}-\yellow{E}) are in far actual distances while intra-patch attention pairs (\red{A}-\red{B}) are in much nearer actual distances in the original video, we propose GRPB to explicitly distinguish these two kinds of attention pairs to avoid confusion of discontinuity between patches and authentic video artifacts. In addition, due to the discontinuity, different mini-patches contain diverse quality information (Fig.~\ref{fig:3}(b)), thus pooling operation before score regression applied in existing methods may confuse the information. To address this issue, we design IP-NLR as a quality-sensitive head, which first regresses the quality scores of mini-patches independently with non-linear layers and pools them after the regression.

In summary, we propose the FrAgment Sample Transformer for VQA (\textbf{FAST-VQA}), with the following contributions:

\begin{enumerate}
\item We propose \frag, a new sampling strategy for VQA that preserves both local quality and unbiased global quality with contextual relations via uniform Grid Mini-patch Sampling (GMS). The \frag~can reduce the complexity of assessing 1080P videos by 97.6\% and enables effective end-to-end training of VQA with quality-retained video samples.

\item We propose the Fragment Attention Network (FANet) to learn the local and contextual quality information from \frag, in which the Gated Relative Position Biases (GRPB) module is proposed to distinguish the intra-patch and cross-patch self-attention and the Intra-Patch Non-Linear Regression (IP-NLR) is proposed for better quality regression from \frag.


\item The proposed FAST-VQA can learn \textit{video-quality-related representations} efficiently through end-to-end training. These quality features help FAST-VQA to be \textbf{10\%} more accurate than the existing state-of-the-art approaches and \textbf{8\%} better than full-resolution Swin-T baseline with fixed recognition features. Through transfer learning, these quality features also significantly improve the best benchmark performance for small VQA datasets.
\end{enumerate}

\section{Related Works}

\paragraph{{Classical VQA Methods}} Classical VQA methods \cite{vbliinds,viideo,tlvqm,videval,rapique,tpqi} handcrafted features to evaluate video quality. Among recent works, TLVQM \cite{tlvqm} uses a combination of spatial high-complexity and temporal low-complexity handcraft features and VIDEVAL \cite{videval} ensembles different handcraft features to model the diverse authentic distortions. However, the reasons affecting the video quality are quite complicated and cannot be well captured with these handcrafted features.


\paragraph{Fixed-feature-based Deep VQA Methods} Due to the extremely high computational cost of deep networks on high resolution videos, existing deep VQA methods train only a feature regression network with fixed deep features. Among them, VSFA \cite{vsfa} uses the features extracted by pre-trained ResNet-50 \cite{he2016residual} from ImageNet-1k \cite{imagenet} and GRU~\cite{gru} for temporal regression. MLSP-FF~\cite{mlsp} also uses heavier Inception-ResNet-V2 \cite{irnv2} for feature extraction. Some methods~\cite{pvq,lsctphiq} use the features extractor pre-trained with IQA datasets~\cite{koniq,paq2piq}. PVQ~\cite{pvq} also extracts features pretrained on action recognition dataset \cite{k400data} for better perception on inter-frame distortion. These methods are limited by their high computational cost on high resolution videos. Additionally, without end-to-end training, fixed features pretrained by other tasks are not optimal for extracting quality-related information, which also limits the accuracy of quality assessment.

\paragraph{VQA Datasets} 

Tab.~\ref{tab:videodatasize} shows common VQA datasets, other video datasets and their sizes. The early VQA datasets~\cite{cvd,qualcomm} are synthesized with specialized distortion and have a very small volume. Some recent in-the-wild VQA datasets like KoNViD-1k~\cite{kv1k}, YouTube-UGC~\cite{ytugc} and LIVE-VQC~\cite{vqc} are still small compared to datasets for other video tasks such as \cite{k400data,activitynet,ava}.
Recently, LSVQ\cite{pvq}, a large-scale VQA dataset with 39,076 videos is publicly available. With end-to-end deep learning of the proposed FAST-VQA, the \textit{video-quality-related} features learnt on large-scale LSVQ dataset can be transferred into smaller VQA datasets to reach better performance.

\begin{table}[]
    \centering
    \vspace{-25pt}
    \setlength\tabcolsep{2.5pt}
    \caption{Common datasets in VQA and other video tasks. Most common VQA datasets are to small (noted in \red{red}) to learn sufficient quality representations independently.}
    \resizebox{0.75\textwidth}{!}{\begin{tabular}{l|c|c|c} \hline
         Dataset & Task & Distortion Type & Size  \\ \hline
         Kinetics-400 \cite{k400data} & Video Recognition & NA & 306,245 \\
         ActivityNet \cite{activitynet} & Video Action Localization & NA & 27,801\\
         AVA \cite{ava} & Atomic Action Detection & NA & 386,000\\\hdashline
         CVD2014 \cite{cvd} & Video Quality Assessment & Synthetic In-capture & \red{234} \\
         KoNViD-1k \cite{kv1k} & Video Quality Assessment & In-the-wild & \red{1,200} \\
         LIVE-VQC \cite{vqc} & Video Quality Assessment & In-the-wild & \red{585} \\
         Youtube-UGC \cite{ytugc} & Video Quality Assessment & In-the-wild & \red{1,147} \\\hdashline
         LSVQ \cite{pvq} & Video Quality Assessment & In-the-wild & 39,076 \\ \hline
    \end{tabular}}
    \label{tab:videodatasize}
    \vspace{-20pt}
\end{table}

\paragraph{Vision Transformers} Vision transformers~\cite{vit,deit,vivit,mvit,swin2d} have shown effective on computer vision tasks. They cut images or videos into non-overlapping patches as input and perform self-attention operations between them. The patch-wise operations in vision transformers naturally distinguish the edges of mini-patches and are suitable for handling with the proposed \frag. 

\begin{figure*}[htbp]
    \centering
    \vspace{-5pt}
    \includegraphics[width=\linewidth]{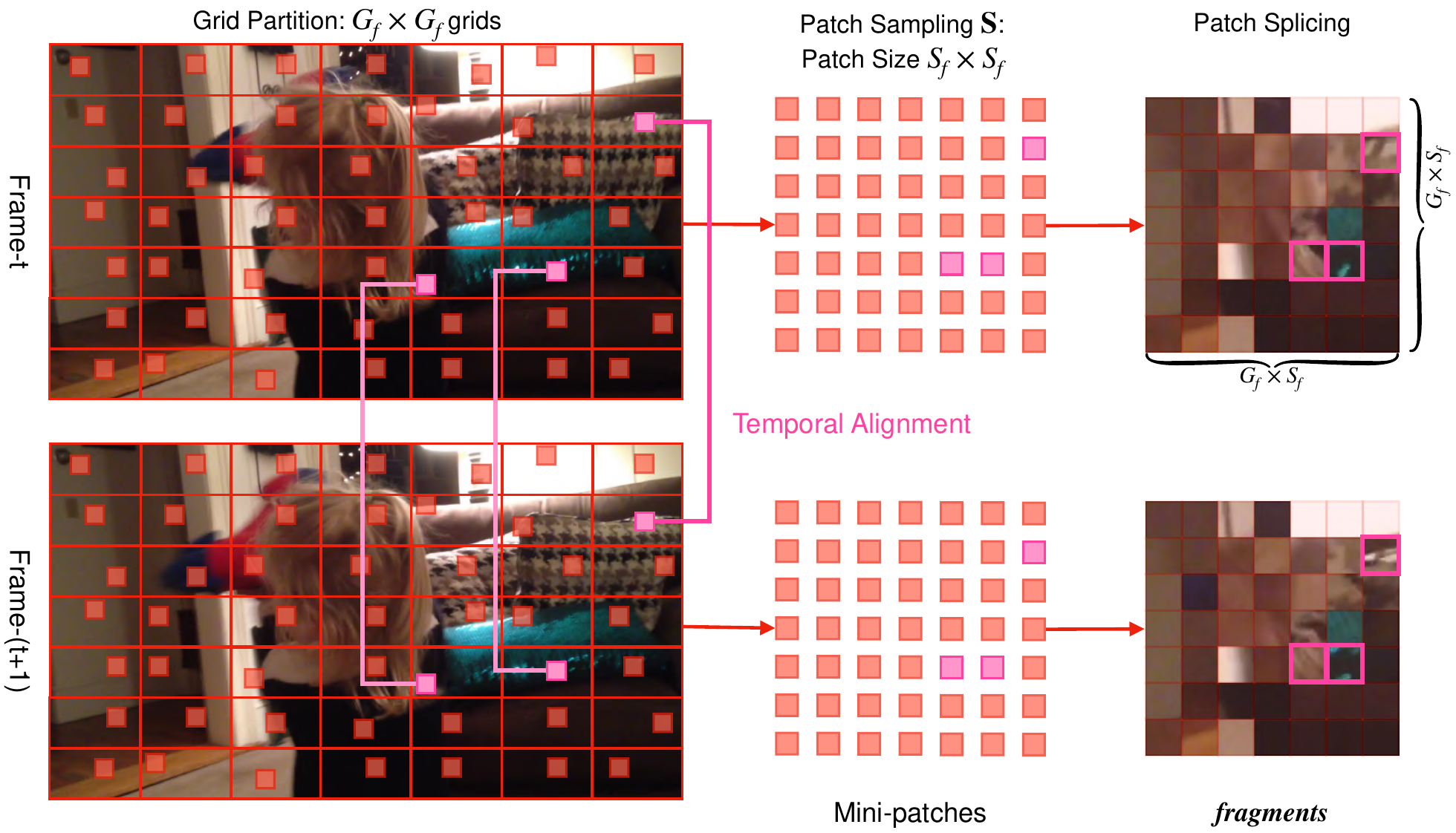}
    \caption{The pipeline for sampling \frag~with Grid Mini-patch Sampling (GMS), including grid partition, patch sampling, patch splicing, and temporal alignment. After GMS, the \frag~are fed into the FANet (Fig.~\ref{fig:5}).}
    \label{fig:4}
    \vspace{-15pt}
\end{figure*}

\section{Approach}
\label{section:method}

In this section, we introduce the full pipeline of the proposed FAST-VQA method. An input video is first sampled into \frag~via Grid Mini-patch Sampling (GMS, Sec.~\ref{section:fragment}). After sampling, the resultant fragments are fed into the Fragment Attention Network (FANet, Sec.~\ref{section:fanet}) to get the final prediction of the video's quality. We introduce both parts in the following subsections.

\subsection{Grid Mini-patch Sampling (GMS)}\label{section:fragment}

To well preserve the original video quality after sampling, we follow several important principles when designing the sampling process for \frag. We will illustrate the process along with these principles below.

\paragraph{Preserving global quality: uniform grid partition.}  To include each region for quality assessment and uniformly assess quality in different areas, we design the grid partition to cut video frames into uniform grids with each grid having the same size (as shown in Fig.~\ref{fig:4}). We cut the $t$-th video frame $\mathcal{V}_t$ into $G_f\times G_f$ uniform grids with the same sizes, denoted as $\mathcal{G}_t=\{g_t^{0,0},..g_t^{i,j},..g_t^{G_f-1,G_f-1}\}$, where $g_t^{i,j}$ denotes the grid 
in the $i$-th row and $j$-th column. The uniform grid partition process is formalized as follows.
\begin{equation}
    g_t^{i,j} = \mathcal{V}_t[\frac{i\times H}{G_f}:\frac{(i+1)\times H}{G_f},\frac{j\times W}{G_f}:\frac{(j+1)\times W}{G_f}]\label{eq:1}
\end{equation}
where $H$ and $W$ denote the height and width of the video frame.

\paragraph{Preserving local quality: raw patch sampling.} To preserve the local textures (\textit{e.g.} blurs, noises, artifacts) that are vital in VQA, we select raw resolution patches without any resizing operations to represent local textural quality in grids. We employ random patch sampling to select one mini-patch $\mathcal{MP}_t^{i,j}$ of size of $S_f \times S_f$ from each grid $g_t^{i,j}$. The patch sampling process is as follows.

\begin{equation}
     \mathcal{MP}_{t}^{i,j} = \mathbf{S}_{t}^{i,j}(g_{t}^{i,j})\label{eq:2}
\end{equation}
where $\mathbf{S}_{t}^{i,j}$ is the patch sampling operation for frame $t$ and grid $i,j$.

\paragraph{Preserving temporal quality: temporal alignment.} It is widely recognized by early works \cite{deepvqa,tlvqm,pvq} that inter-frame temporal variations are influential to video qualities. To retain the raw temporal variations in videos (with $T$ frames), we strictly align the sample areas during patch sampling operations $\mathbf{S}$ in different frames, as the following constraint shows.

\begin{equation}
     \mathbf{S}_{t}^{i,j} = \mathbf{S}_{\hat{t}}^{i,j}~~~~~~\forall~0\leq t,\hat{t}<T,~ 0\leq i, j < G_f
\end{equation}

\paragraph{Preserving contextual relations: patch splicing.} Existing works~\cite{sfa,vsfa,mlsp} have shown that the global scene information and contextual information affects quality predictions. To keep the global scene information of the original videos, we keep the contextual relations of mini-patches by splicing them into their original positions, as the following equation shows:

\begin{equation}
\begin{aligned}
        \mathcal{F}_t^{i,j}&= \mathcal{F}_t[i\times S_f:(i+1)\times S_f, j\times S_f:(j+1)\times S_f] \\&= \mathcal{MP}_t^{i,j},~~~~~~~~~~~~~~~~~~~~ 0\leq i, j < G_f \label{eq:3} 
\end{aligned}
\end{equation}
where $\mathcal{F}$ denote the spliced and temporally aligned mini-patches after the Grid Mini-patch Sampling (GMS) pipeline, named as \frag.

\subsection{Fragment Attention Network (FANet)}\label{section:fanet}

\paragraph{The Overall Framework.} Fig.~\ref{fig:5} shows the overall framework of FANet. It uses a Swin-T with four hierarchical self-attention layers as backbone. We also design the following modules to adapt it to fragments well.
\paragraph{Gated Relative Position Biases.} Swin-T adds relative position bias (RPB) that uses learnable Relative Bias Table ($\mathbf{T}$) to represent the relative positions of pixels in attention pairs ($QK^T$). For \frag, however, as discussed in Fig.~\ref{fig:3}(a), the cross-patch pairs have much large actual distances than intra-patch pairs and should not be modeled with the same bias table. Therefore, we propose the gated relative position biases (GRPB, Fig.~\ref{fig:5}(b)) that uses learnable real position bias table ($\mathbf{T}^\text{real}$) and pseudo position bias table ($\mathbf{T}^\text{pseudo}$) to replace $\mathbf{T}$. 
The mechanisms of them are the same as $\mathbf{T}$ but they are learnt separately and used for intra-patch and cross-patch attention pairs respectively.
Denote $\mathbf{G}$ as the intra-patch gate ($\mathbf{G}_{i,j}=1$ if $i,j$ are in the same mini-patch else $\mathbf{G}_{i,j}=0$), the self-attention matrix ($M_A$) with GRPB is calculated as:


\begin{figure}[]
    \centering
    \includegraphics[width=0.96\linewidth]{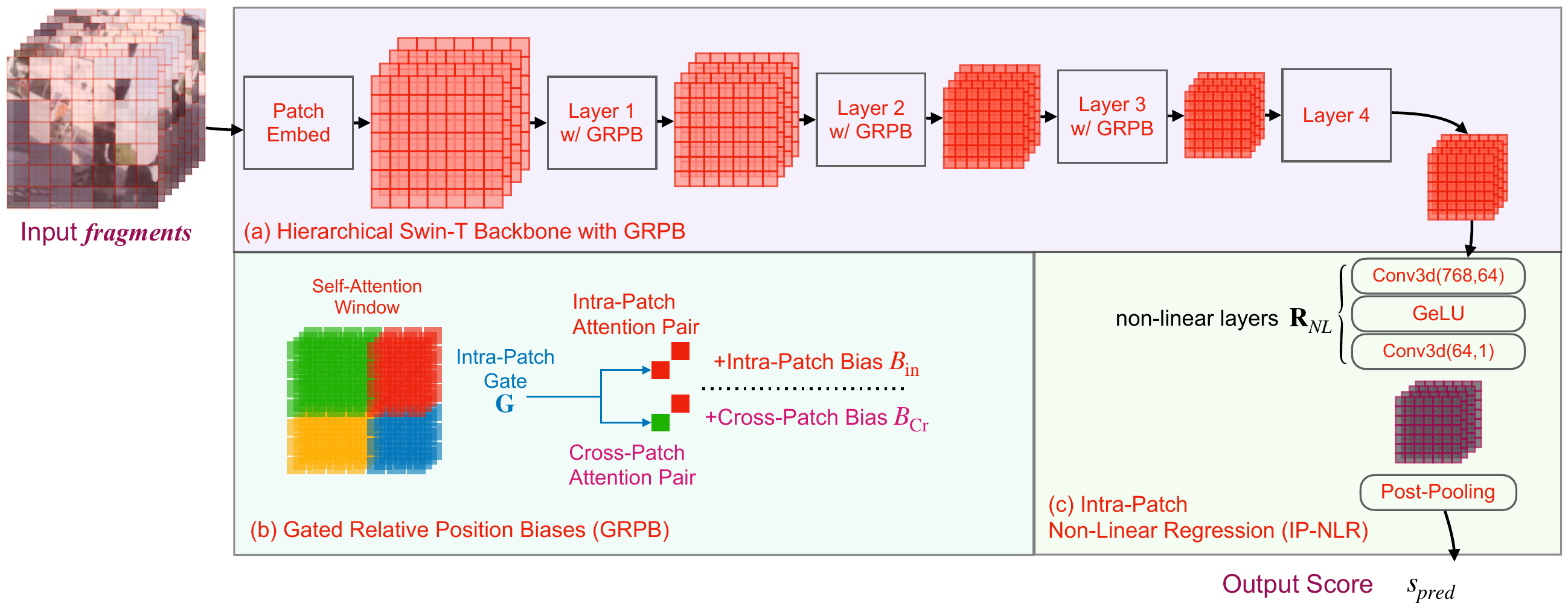}
    \vspace{-10pt}
    \caption{The overall framework for FANet, including the Gated Relative Position Biases (GRPB) and Intra-Patch Non-Linear Regression (IP-NLR) modules. The input \frag~come from Grid Mini-patch Sampling (Fig.~\ref{fig:4}).}
    \label{fig:5}
    \vspace{-18pt}
\end{figure}

\vspace{-10pt}
\begin{align}
    B_{\text{In}, (i,j)} &= \mathbf{T}^{\text{real}}_{\mathrm{FRP}(i,j)};
    B_{\text{Cr}, (i,j)} = \mathbf{T}^{\text{pseudo}}_{\mathrm{FRP}(i,j)}  \\
    M_A &= QK^T + \mathbf{G} \otimes B_{\text{In}} + (\mathbf{1}-\mathbf{G}) \otimes B_\text{{Cr}} 
\end{align}
where $\mathrm{FRP}(i,j)$ is the relative position of pair $(i,j)$ in \frag.

\paragraph{Intra-Patch Non-Linear Regression.} As illustrated in Fig.~\ref{fig:3}(b), different mini-patches have diverse qualities due to discontinuity between them. If we pool features from different patches before regression, the quality representations of mini-patches will be confused with each other. To avoid this problem, we design the Intra-Patch Non-Linear Regression (IP-NLR, Fig.~\ref{fig:5}(c)) to regress the features via non-linear layers ($\mathbf{R}_\textit{NL}$) first, and perform pooling following the regression. Denote features as $f$, output score as $s_{pred}$, pooling operation as $Pool(\cdot)$, the IP-NLR can be expressed as follows:

\vspace{-10pt}
\begin{align}
    s_{pred} &= Pool(\mathbf{R}_\textit{NL}(f))
\end{align}

\section{Experiments}

In the experiment part, we conduct several experiments to evaluate and analyze the performance of the proposed FAST-VQA model.

\subsection{Evaluation Setup}

\paragraph{Implementation Details} We use the Swin-T~\cite{swin3d} pretrained on Kinetics-400~\cite{k400data} dataset to initialize the backbone in FANet. As Tab.~\ref{tab:impdetail} shows, we implement two sampling densities for \frag:  FAST-VQA (normal density) and FAST-VQA-M (lower density \& higher efficiency), and accomodate window sizes in FANet to the input sizes.  Without special notes, all ablation studies are on variants of FAST-VQA. We use PLCC (Pearson linear correlation coef.) and SRCC (Spearman rank correlation coef.) as metrics and use differentiable PLCC loss $l=\frac{(1-\mathrm{PLCC}(s_{pred},s_{gt}))}{2}$ as loss function. We set the training batch size as 16.

\vspace{-25pt}
\begin{table}[]
    \centering
    \setlength\tabcolsep{5pt}
    \caption{Comparison of FAST-VQA and FAST-VQA-M with lower sampling density.}
    \resizebox{\linewidth}{!}{
    \begin{tabular}{l|c|c|c|c|c|c} \hline
    \makecell[c]{Methods}&\makecell{Number of\\ Frames ($T$)}&\makecell[c]{Patch Size\\ ($S_f$)}&\makecell[c]{Number of \\Grids ($G_f$)}&\makecell[c]{Window Size \\ in FANet}&FLOPs&Parameters\\\hline
    \textbf{FAST-VQA}&32&32&7&(8,7,7)&279G&27.7M\\
    \textbf{FAST-VQA-M}&16&32&4&(4,4,4)&46G&27.5M\\
 \hline
    \end{tabular}
    }
    \label{tab:impdetail}
    \vspace{-14pt}
\end{table}

\paragraph{Training \& Benchmark Sets} We use the large-scale LSVQ{$_\text{train}$}\cite{pvq} dataset with 28,056 videos for training FAST-VQA. For evaluation, we choose 4 testing sets to test the model trained on LSVQ. The first two sets, LSVQ$_\text{test}$ and LSVQ$_\text{1080p}$ are official intra-dataset test subsets for LSVQ, while the LSVQ$_\text{test}$ consists of 7,400 various resolution videos from 240P to 720P, and LSVQ$_\text{1080p}$ consists of 3,600 1080P high resolution videos. We also evaluate the generalization ability of FAST-VQA on cross-dataset evaluations on KoNViD-1k \cite{kv1k} and LIVE-VQC \cite{vqc}, two widely-recognized in-the-wild VQA benchmark datasets. 

\subsection{Benchmark Results}

\begin{table}[]
\footnotesize
\vspace{-25pt}
\caption{Comparison with existing methods (classical and deep) and our baseline (Full-res Swin-T \textit{features}). The 1st/2nd best scores are colored in \textbf{\red{red}} and \blue{blue}, respectively.} \label{table:peer}
\vspace{-3mm}
\setlength\tabcolsep{5.4pt}
\renewcommand\arraystretch{1.25}
\footnotesize
\center
\resizebox{1.\textwidth}{!}{\begin{tabular}{l:l|cc|cc|cc|cc}
\hline
\multicolumn{2}{l|}{Type/} & \multicolumn{4}{c|}{Intra-dataset Test Sets}        &  \multicolumn{4}{c}{Cross-dataset Test Sets}            \\ \hdashline
\multicolumn{2}{l|}{\textbf{Testing Set}/}         & \multicolumn{2}{c|}{\textbf{LSVQ$_\text{test}$}}   & \multicolumn{2}{c|}{\textbf{LSVQ$_\text{1080p}$}}        &  \multicolumn{2}{c|}{\textbf{KoNViD-1k}}  & \multicolumn{2}{c}{\textbf{LIVE-VQC}}             \\ \hline
Groups~~~ & ~Methods                   & SRCC & PLCC    & SRCC & PLCC           & SRCC & PLCC                      & SRCC & PLCC                               \\ \hline 
\multirow{3}{0pt}{{{Existing Classical}}} &~BRISQUE\cite{brisque}    & 0.569 &  0.576  & 0.497 & 0.531     & 0.646 &     0.647                   & 0.524 &  0.536 \\
&~TLVQM\cite{tlvqm}        & 0.772 &  0.774  & 0.589 & 0.616     & 0.732 &     0.724                   & 0.670 &  0.691 \\
&~VIDEVAL\cite{videval}      & 0.794 &  0.783  & 0.545 & 0.554     & 0.751 &     0.741                   & 0.630 &  0.640 \\\hdashline    
\multirow{3}{0pt}{{{Existing Deep}}} &~VSFA\cite{vsfa}          & 0.801 &  0.796  & 0.675 & 0.704     & 0.784 &     0.794                   & 0.734 &  0.772 \\

&~PVQ$_\text{wo/ patch}$\cite{pvq}   & 0.814 &  0.816  & 0.686 &   0.708         & 0.781 &     0.781                   & 0.747 &  0.776                               \\ 
&~PVQ$_\text{w/ patch}$\cite{pvq}   & 0.827 &  0.828 & 0.711 &  0.739     & 0.791 &     0.795  & 0.770 &  0.807   \\ \hdashline
\multicolumn{2}{l|}{Full-res Swin-T\cite{swin3d} \textit{features}} &  0.835 & 0.833 & 0.739 & 0.753 & 0.825 & 0.828         &  \blue{0.794} & 0.809 \\ \hline
\multicolumn{2}{l|}{\textbf{FAST-VQA-M} (Ours)} &  \blue{0.852} & \blue{0.854} & \blue{0.739} & \blue{0.773} & \blue{0.841} & \blue{0.832} & {0.788} & \blue{0.810} \\ \hline
\multicolumn{2}{l|}{\textbf{FAST-VQA} (Ours)} &  \textbf{\red{0.876}} & \textbf{\red{0.877}}  & \textbf{\red{0.779}} & \textbf{\red{0.814}} & \textbf{\red{0.859}} & \textbf{\red{0.855}}& \textbf{\red{0.823}} & \textbf{\red{0.844}}  \\ \hline
\multicolumn{2}{l|}{\textit{Improvement} to PVQ$_\text{w/ patch}$} &  \textit{+6\%} & \textit{+6\%} & \textit{+10\%} & \textit{+10\%} & \textit{+9\%} & \textit{+8\%} & \textit{+7\%} & \textit{+5\%}  \\ \hline
\end{tabular}}
\vspace{-4mm}

\end{table}

In Tab.~\ref{table:peer}, we compare with existing classical and deep VQA methods and our baseline, the full-resolution Swin-T with feature regression instead of end-to-end training (denoted as `Full-res Swin-T \textit{features}'). With its video-quality-related representations, FAST-VQA achieves at most 10\% improvement to PVQ, the existing state-of-the-art on LSVQ$_\text{1080p}$. Even the efficient version FAST-VQA-M can outperform existing state-of-the-art. FAST-VQA also shows significant improvement to its fixed-feature-based baseline with the same backbone, demonstrating that the proposed new {{quality-retained sampling}} with {{end-to-end training}} scheme for VQA is not only much more efficient (with only 2.36\% FLOPs required on 1080P videos) but also notably more accurate (with 8.10\% improvement on PLCC metric for LSVQ$_\text{1080p}$) than the 
existing fixed-feature-based paradigm.

\subsection{Efficiency of FAST-VQA}
\label{sec:eff}

\begin{figure}[t]
    \centering
    \includegraphics[width=\linewidth]{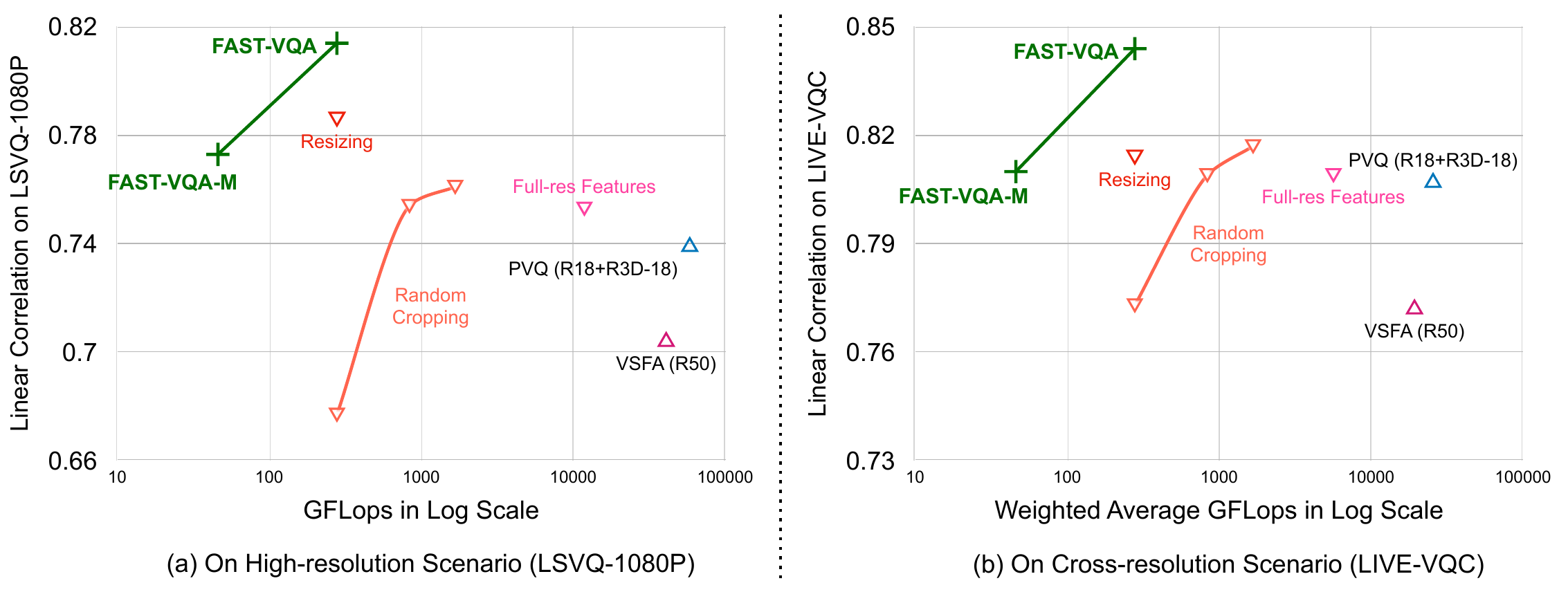}
    \vspace{-15pt}
    \caption{The Performance-FLOPs curve of proposed \green{\textbf{FAST-VQA}} and baseline methods.}
    \label{fig:a1}
\end{figure}

\begin{table*}[t]
\centering
    \caption{FLOPs and running time (on GPU/CPU, average of ten runs) comparison of FAST-VQA, state-of-the-art methods and our baseline on different resolutions. We \textbf{boldface} FLOPs $\leq$ 500G and running time $\leq$ 1s.} \label{tab:efficiency}
    \vspace{3pt}
    \setlength\tabcolsep{6pt}
    \renewcommand\arraystretch{1.15}
\resizebox{.92\textwidth}{!}{
    \begin{tabular}{l|c:c|c:c|c:c}
    \hline
    {} & \multicolumn{2}{c|}{540P} & \multicolumn{2}{c|}{720P} & \multicolumn{2}{c}{1080P}\\
    \cline{2-7}
    {Method} & FLOPs(G) & Time(s)  & FLOPs(G) & Time(s) & FLOPs(G) & Time(s)\\
    \hline
     VSFA\cite{vsfa} & 10249\red{$_{36.7\times}$} & 2.603/92.761 & 18184\red{$_{65.2\times}$} & 3.571/134.9 & 40919\red{$_{147\times}$} & 11.14/465.6 \\ 
     PVQ\cite{pvq} & 14646\red{$_{52.5\times}$} & 3.091/97.85 & 22029\red{$_{79.0\times}$} & 4.143/144.6 & 58501\red{$_{210\times}$} & 13.79/538.4 \\ \hdashline
     Full-res Swin-T\cite{swin3d} \textit{feat.} & 3032\red{$_{10.9\times}$} & 3.226/102.0 & 5357\red{$_{19.2\times}$} & 5.049/166.2 & 11852\red{$_{42.5\times}$} & 8.753/234.9 \\ \hline
        \textbf{FAST-VQA} (Ours) & {\textbf{279}{$_{1\times}$}}& \textbfg{0.044}/9.019 & {\textbf{279}{$_{1\times}$}}& \textbfg{0.043}/9.530 & {\textbf{279}{$_{1\times}$}} & \textbfg{0.045}/9.142 \\ 
    \textbf{FAST-VQA-M} (Ours) & \textbfg{46}\green{$_{0.165\times}$}& \textbfg{0.019/0.729} & \textbfg{46}\green{$_{0.165\times}$}& \textbfg{0.019/0.613} & \textbfg{46}\green{$_{0.165\times}$} & \textbfg{0.019/0.714} \\ \hline
    \end{tabular}

}
\end{table*}

To demonstrate the efficiency of FAST-VQA, we compare the FLOPs and running times on CPU/GPU (average of ten runs per sample) of the proposed FAST-VQA with existing deep VQA approaches on different resolutions, see Tab.~\ref{tab:efficiency}. We also draw the performance-FLOPs curve on LSVQ$_{1080p}$ and LIVE-VQC in Fig.~\ref{fig:a1}. As we can see, FAST-VQA reduces up to $210\times$ FLOPs and $247\times$ running time than PVQ while obtaining notably better performance. The more efficient version, FAST-VQA-M, only requires $1/1273$ FLOPs of PVQ and $1/258$ FLOPs of our full-resolution baseline while still achieving slightly better performance. Moreover, FAST-VQA (especially FAST-VQA-M) also runs very fast even on CPU, which reduces the hardware requirements for the applications of deep 
VQA methods. All these comparisons show the unprecedented efficiency of proposed FAST-VQA. \footnote{Also, RAPIQUE\cite{rapique} can also infer rapidly on CPU that requires \textbf{17.3s} for 1080P videos. Yet, it is not compatible with GPU Inference due to its handcrafted branch.}

\subsection{Transfer Learning with Video-quality-related Representations} 

\vspace{-5pt}
\begin{table}[]
\footnotesize
\caption{The finetune results on LIVE-VQC, KoNViD, CVD2014 and YouTube-UGC datasets, compared with existing classical and fixed-backbone deep VQA methods, and ensemble approaches of classical (C) and deep (D) branches.} \label{table:vqc}
\setlength\tabcolsep{5.5pt}
\renewcommand\arraystretch{1.25}
\footnotesize
\centering
\resizebox{1.\textwidth}{!}{\begin{tabular}{l:l|cc|cc|cc|cc|cc}
\hline
\multicolumn{2}{l|}{\textbf{Finetune Dataset}/}         & \multicolumn{2}{c|}{LIVE-VQC}   & \multicolumn{2}{c|}{KoNViD-1k}        &  \multicolumn{2}{c|}{CVD2014}   &  \multicolumn{2}{c|}{LIVE-Qualcomm}  & \multicolumn{2}{c}{YouTube-UGC}             \\ \hline
Groups~~~ &~Methods                 & SRCC & PLCC    & SRCC & PLCC           & SRCC & PLCC          &SRCC&PLCC            &SRCC& PLCC                               \\ \hline 
\multirow{3}{0pt}{{{Existing Classical}}} &~TLVQM\cite{tlvqm}        & 0.799 &  0.803  & 0.773 & 0.768     & 0.83 &    0.85               & 0.77 & 0.81 & 0.669 &  0.659 \\

&~VIDEVAL\cite{videval}      & 0.752 &  0.751  & 0.783 & 0.780     & NA &     NA    & NA & NA               & 0.779 &  0.773 \\
&~RAPIQUE\cite{rapique}      & 0.755 &  0.786  & 0.803 & 0.817     & NA &     NA  & NA & NA                 & 0.759 &  0.768 \\\hdashline 
\multirow{4}{0pt}{{{Existing \textbf{Fixed} Deep}}} &~VSFA\cite{vsfa}          & 0.773 &  0.795  & 0.773 & 0.775     & 0.870 &     0.868 & 0.737 & 0.732 & 0.724 &  0.743\\
&~PVQ\cite{pvq}   & \blue{0.827} &  \blue{0.837}  & 0.791 &   0.786       & NA & NA  & NA &     NA                   & NA &  NA\\ 
&~GST-VQA\cite{gstvqa}  & NA &  NA  & 0.814 &   0.825         & 0.831 & 0.844  & 0.801 &  0.825  & NA & NA\\ 
&~CoINVQ\cite{rfugc} & NA &  NA & 0.767 &  0.764  & NA & NA &  NA & NA   & \blue{0.816} &     {0.802}  \\ \hdashline
\multirow{2}{0pt}{{{Ensemble C+D}}} & CNN+TLVQM\cite{cnntlvqm}        & 0.825 & 0.834 & 0.816 & 0.818 & 0.863 & 0.880  & \blue{0.810} & 0.833 & NA & NA \\
& CNN+VIDEVAL\cite{videval}        & 0.785 & 0.810 & 0.815 & 0.817 & NA & NA  & NA & NA & 0.808 & \blue{0.803} \\\hdashline
\multicolumn{2}{l|}{Full-res Swin-T\cite{swin3d} \textit{features}} & 0.799 & 0.808 & 0.841 & 0.838 & 0.868 & 0.870 & 0.788 & 0.803 & 0.798 & 0.796 \\ \hline

\multicolumn{2}{l|}{{FAST-VQA-M} (Ours)} & 0.803 & 0.828 & \blue{0.873} & \blue{0.872} & \blue{0.877} & \blue{0.892} & 0.804 & \blue{0.838} & 0.768 & 0.765\\ \hline
\multicolumn{2}{l|}{{FAST-VQA} \textit{w/o VQ-representations} (Ours)} & 0.765 & 0.782 & 0.842 & 0.844 & 0.871 & 0.888 & 0.756 & 0.778 & 0.794 & 0.784 \\ \hdashline
\multicolumn{2}{l|}{\textbf{FAST-VQA} (ours)} &  \textbf{\red{0.849}} & \textbf{\red{0.865}} & \textbf{\red{0.891}} & \textbf{\red{0.892}} & \textbf{\red{0.891}} & \textbf{\red{0.903}} & \textbf{\red{0.819}} & \textbf{\red{0.851}} & \textbf{\red{0.855}} & \textbf{\red{0.852}}  \\ \hline
\multicolumn{2}{l|}{Improvements led by \textit{VQ-representations}} & \red{+11.0\%} & \red{+10.6\%}  & \red{+5.8\%}  &\red{+5.7\%}  & \red{+2.3\%}  & \red{+1.7\%}  & \red{+8.3\%}  & \red{+9.4\%}  & \red{+7.7\%}  & \red{+8.7\%}  \\  \hline
\end{tabular}}
\vspace{-12pt}
\end{table}

FAST-VQA also makes the pretrain-finetune scheme on VQA possible with affordable computation resources. With FAST-VQA, we can pretrain with large VQA datasets in end-to-end manner to learn quality related features, and then transfer to specific VQA scenarios where only small datasets are available. Note that this manner is not applicable to current methods due to their high computational load (as discussed in Sec.~\ref{sec:eff}). We use LSVQ as the large dataset and choose four small datasets representing diverse scenarios, including LIVE-VQC (real-world mobile photography, 240P-1080P), KoNViD-1k (various contents collected online, all 540P), CVD2014 (synthetic in-capture distortions, 480P-720P), LIVE-Qualcomm (selected types of distortions, all 1080P) and YouTube-UGC (user-generated contents, including computer graphic contents, 360P-2160P\footnote{Due to privacy reasons, the current public version of YouTube-UGC is incomplete and only with 1147 videos. The peer comparison is only for reference.}). We divide each dataset into random splits for 10 times and report the average result on the test splits. As Tab.~\ref{table:vqc} shows, with video-quality-related representations, the proposed FAST-VQA outperforms the existing state-of-the-arts on all these scenarios while obtaining much higher efficiency. Note that YouTube-UGC contains 4K(2160P) videos but FAST-VQA still performs well. Even without video-quality-related representations, FAST-VQA also still achieves competitive performance, while these features steadily improve the performance. It implies that the pretrained FAST-VQA can be able to serve as a strong backbone that boost further downstream tasks related to video quality.

\subsection{Ablation Studies on \textit{fragments}}

For the first part of ablation studies, we prove the effectiveness of \frag~by comparing with other common sampling approaches and different variants of fragments (Tab.~\ref{tab:gmstfa}). We keep the FANet structure fixed during this part.

\begin{table}[]
\footnotesize
\vspace{-6pt}
\caption{Ablation study on \frag: comparison with resizing, cropping (Group 1) and different variants for fragments (Group 2).} 
\setlength\tabcolsep{5pt}
\renewcommand\arraystretch{1.15}
\footnotesize
\centering
\label{tab:resizecrop}
\resizebox{.8\textwidth}{!}{\begin{tabular}{l|cc|cc|cc|cc}
\hline
\textbf{Testing Set}/         & \multicolumn{2}{c|}{\textbf{LSVQ$_\text{test}$}}   & \multicolumn{2}{c|}{\textbf{LSVQ$_\text{1080p}$}}        &  \multicolumn{2}{c|}{\textbf{KoNViD-1k}}  & \multicolumn{2}{c}{\textbf{LIVE-VQC}}             \\ \cline{2-9}
Methods/Metric                    & SRCC & PLCC    & SRCC & PLCC           & SRCC & PLCC                      & SRCC & PLCC     \\ \hline     
\multicolumn{9}{l}{{Group 1: Naive Sampling Approaches}} \\ \hdashline
\textit{bilinear resizing} &  0.857 & 0.859 & 0.752 & 0.786 & 0.841 & 0.840         &  0.772 & 0.814 \\ \hdashline
\textit{random cropping} &  0.807 & 0.812 & 0.643 & 0.677 & 0.734 & 0.776         &  0.740 & 0.773 \\
- test with 3 crops &  0.838 & 0.835 & 0.727 & 0.754 & 0.841 & 0.827         &  0.785 & 0.809 \\
- test with 6 crops &  0.843 & 0.844 & 0.734 & 0.761 & 0.845 & 0.834         &  0.796 & 0.817 \\ 
\hline
\multicolumn{9}{l}{{Group 2: Variants of \frag}}    \\ \hdashline                
\textit{random mini-patches} &  0.857 & 0.861 & 0.754 & 0.790 & 0.844 & 0.845 &  0.792 & 0.818 \\ 
\textit{shuffled mini-patches} &  0.858 & 0.863 & 0.761 & 0.799 & 0.849 & 0.847 &  0.796 & 0.821 \\ \hdashline
\textit{w/o} temporal alignment &  0.850 & 0.853 & 0.736 & 0.779 & 0.823 & 0.816         &  0.764 & 0.802 \\
\hline
\textit{\textbf{fragments}} (ours) &  \textbf{\red{0.876}} & \textbf{\red{0.877}}  & \textbf{\red{0.779}} & \textbf{\red{0.814}} & \textbf{\red{0.859}} & \textbf{\red{0.855}}& \textbf{\red{0.823}} & \textbf{\red{0.844}} \\ \hline
\end{tabular}}
\label{tab:gmstfa}
\vspace{-15pt}
\end{table}

\paragraph{Comparing with resizing/cropping} In Group 1 of Tab.~\ref{tab:resizecrop}, we compare the proposed fragments with two common sampling approaches: \textit{bilinear resizing} and \textit{random cropping}. The proposed \textit{fragments} are notably better than bilinear resizing on \textbf{high-resolution} (LSVQ$_\text{1080p}$) (+4\%) and \textbf{cross-resolution} (LIVE-VQC) scenarios (+4\%). Fragments still lead to non-trivial 2\% improvements than resizing on lower-resolution scenarios where the problems of resizing is not that severe. This proves that keeping local textures is vital for VQA. Fragments also largely outperform single random crop as well as ensemble of multiple crops, suggesting that retaining the uniform global quality is also critical to VQA.

\paragraph{Comparing with variants of fragments} We also compare with three variants of \frag~in Tab.~\ref{tab:gmstfa}, Group 2. We prove the effectiveness of uniform grid partition by comparing with \textit{random mini-patches} (ignore grids while sampling), and the importance of retaining contextual relations by comparing with \textit{shuffled mini-patches}. Fragments show notable improvements than both variants. Moreover, the proposed fragments show much better performance than the variant \textit{without} temporal alignment especially on high resolution videos, suggesting that preserving the inter-frame temporal variations is necessary for fragments. 

\begin{table}
\setlength\tabcolsep{5pt}
\renewcommand\arraystretch{1.15}
\footnotesize
\vspace{-22pt}
\caption{Ablation study on FANet design: the effects for GRPB and IP-NLR modules.} 
\centering
\resizebox{.8\textwidth}{!}{\begin{tabular}{l|cc|cc|cc|cc}
\hline
\textbf{Testing Set}/         & \multicolumn{2}{c|}{\textbf{LSVQ$_\text{test}$}}   & \multicolumn{2}{c|}{\textbf{LSVQ$_\text{1080p}$}}        &  \multicolumn{2}{c|}{\textbf{KoNViD-1k}}  & \multicolumn{2}{c}{\textbf{LIVE-VQC}}             \\ \cline{2-9}
Variants/Metric                   & SRCC & PLCC    & SRCC & PLCC           & SRCC & PLCC                      & SRCC & PLCC     \\ \hline                
\textit{w/o} GRPB &  0.873 & 0.872 & 0.769 & 0.805 & 0.854 & 0.853 &  0.808 & 0.832 \\ 
\textit{semi}-GRPB on Layer 1/2 &  0.873 & 0.875 & 0.772 & 0.809 & 0.856 & 0.851 &  0.812 & 0.838 \\ \hdashline
\textit{linear} Regression &  0.872 & 0.873 & 0.768 & 0.803 & 0.847 & 0.849         &  0.810 & 0.835 \\
\textit{PrePool non-linear} Regression &  0.873 & 0.874 & 0.771 & 0.805 & 0.851 & 0.850         &  0.813 & 0.834 \\\hline
\textbf{FANet} (ours) &  \textbf{\red{0.876}} & \textbf{\red{0.877}}  & \textbf{\red{0.779}} & \textbf{\red{0.814}} & \textbf{\red{0.859}} & \textbf{\red{0.855}}& \textbf{\red{0.823}} & \textbf{\red{0.844}}  \\ \hline
\end{tabular}}
\label{tab:netdesign}
\vspace{-22pt}
\end{table}

\subsection{Ablation Studies on FANet}

\paragraph{Effects of GRPB and IP-NLR} In the second part of ablation studies, we analyze the effects of two important designs in FANet: the proposed Gated Relative Position Biases (GRPB) and Intra-Patch Non-Linear Regression (IP-NLR) VQA Head as in Tab.~\ref{tab:netdesign}. We compare the IP-NLR with two variants: the linear regression layer and the non-linear regression layers with pooling before regression (\textit{PrePool}). Both modules lead to non-negligible improvements especially on high-resolution (LSVQ$_\text{1080p}$) or cross-resolution (LIVE-VQC) scenarios. As  the discontinuity between mini-patches is more obvious in high-resolution videos, this result suggests that the corrected position biases and regression head are helpful on solving the problems caused by such discontinuity.

\subsection{Reliability and Robustness Analyses}

As FAST-VQA is based on samples rather than original videos while a single sample for \frag~only keeps 2.4\% spatial information in 1080P videos, it is important to analyze the reliability and robustness of FAST-VQA predictions.

\paragraph{Reliability of Single Sampling.} We measure the reliability of single sampling in FAST-VQA by two metrics: 1) the assessment stability of different single samplings on the same video; 2) the relative accuracy of single sampling compared with multiple sample ensemble. As shown in Tab.~\ref{tab:stability}, the normalized \textit{std. dev.} of different sampling on a same video is only around 0.01, which means the sampled fragments are enough to make very stable predictions. Compared with 6-sample ensemble, sampling only once can already be 99.40\% as accurate even on the pure high-resolution test set (LSVQ$_\text{1080P}$). They prove that a single sample of \frag~is enough stable and reliable for quality assessment even though only a small proportion of information is kept during sampling.

\begin{table}
\center
\setlength\tabcolsep{6pt}
\renewcommand\arraystretch{1.15}
\footnotesize
\vspace{-18pt}
\caption{Assessment stability and relative accuracy of single sampling of \frag.} 
\resizebox{.96\textwidth}{!}{\begin{tabular}{l|c|c|c|c}
\hline
\textbf{Testing Set}/         & \multicolumn{1}{|c|}{\textbf{LSVQ$_\text{test}$}}   & \multicolumn{1}{|c|}{\textbf{LSVQ$_\text{1080p}$}}        &  \multicolumn{1}{|c|}{\textbf{KoNViD-1k}}  & \multicolumn{1}{|c}{\textbf{LIVE-VQC}}             \\ \cline{2-5}
Score Range & 0-100 & 0-100 & 1-5 & 0-100 \\ \hline
\textit{std. dev.} of Single Samplings  & 0.65 & 0.79 & 0.046 & 1.07 \\ 
Normalized \textit{std. dev.} & 0.0065 & 0.0079 & 0.0115 & 0.0107 \\ \hline
Relative Pair Accuracy compared with 6-samples & 99.59\% & 99.40\% & 99.45\% & 99.52\%  \\ \hline
\end{tabular}}
\label{tab:stability}
\vspace{-20pt}
\end{table}

\paragraph{Robustness on Different Resolutions}
    To analyze the robustness of FAST-VQA on different resolutions, we divide the cross-resolution VQA benchmark set LIVE-VQC into three resolution groups: (A) 1080P (110 videos); (B) 720P (316 videos); (C) $\leq$540P (159 videos) to see the performance of FAST-VQA on different resolutions, compared with several variants. As the results shown in Tab.~\ref{tab:resolution}, the proposed FAST-VQA shows good performance ($\geq0.80$ SRCC\&PLCC) on all resolution groups and most superior improvement than other variants on Group (A) with 1080P high-resolution videos, proving that FAST-VQA is robust and reliable on different resolutions of videos.

\begin{table}
\center
\setlength\tabcolsep{5pt}
\renewcommand\arraystretch{1.15}
\footnotesize
\vspace{-16pt}
\caption{Performance comparison on different resolution groups of LIVE-VQC dataset.} 
\resizebox{.8\textwidth}{!}{\begin{tabular}{l|ccc|ccc|ccc}
\hline
\textbf{Resolution}     & \multicolumn{3}{c|}{(A): 1080P}   & \multicolumn{3}{c|}{(B): 720P}        &  \multicolumn{3}{|c}{(C): $\leq$540P}    \\ \cline{2-10}
Variants                    & SRCC & PLCC & KRCC    & SRCC & PLCC  & KRCC         & SRCC & PLCC                & KRCC     \\ \hline    
\textit{Full-res} Swin \textit{features} (Baseline) & 0.771 & 0.774 & 0.584 & 0.796 & 0.811 & 0.602 & 0.810 & 0.853 & 0.625 \\ \hdashline
\textit{bilinear resizing} (Sampling Variant) & 0.758 & 0.773 & 0.573 & 0.790 & 0.822 & 0.599 & 0.835 & 0.878 & 0.650 \\ \hdashline
\textit{random cropping} (Sampling Variant) & 0.765 & 0.768 & 0.565 & 0.774 & 0.787 &  0.581 & 0.730 & 0.809 & 0.535 \\  \hdashline

\textit{w/o} GRPB (FANet Variant) & 0.796 & 0.785 & 0.598 & 0.802 & 0.820 & 0.608 & 0.834 & 0.883 & 0.649 \\ \hline
\textbf{FAST-VQA} (Ours) & \bred{0.807} & \bred{0.806} & \bred{0.610} & \bred{0.803} & \bred{0.825} & \bred{0.610} & \bred{0.840} & \bred{0.885} & \bred{0.654} \\ \hline

\end{tabular}}
\label{tab:resolution}
\vspace{-25pt}
\end{table}

\begin{figure}[]
    \centering
    \includegraphics[width=0.88\linewidth]{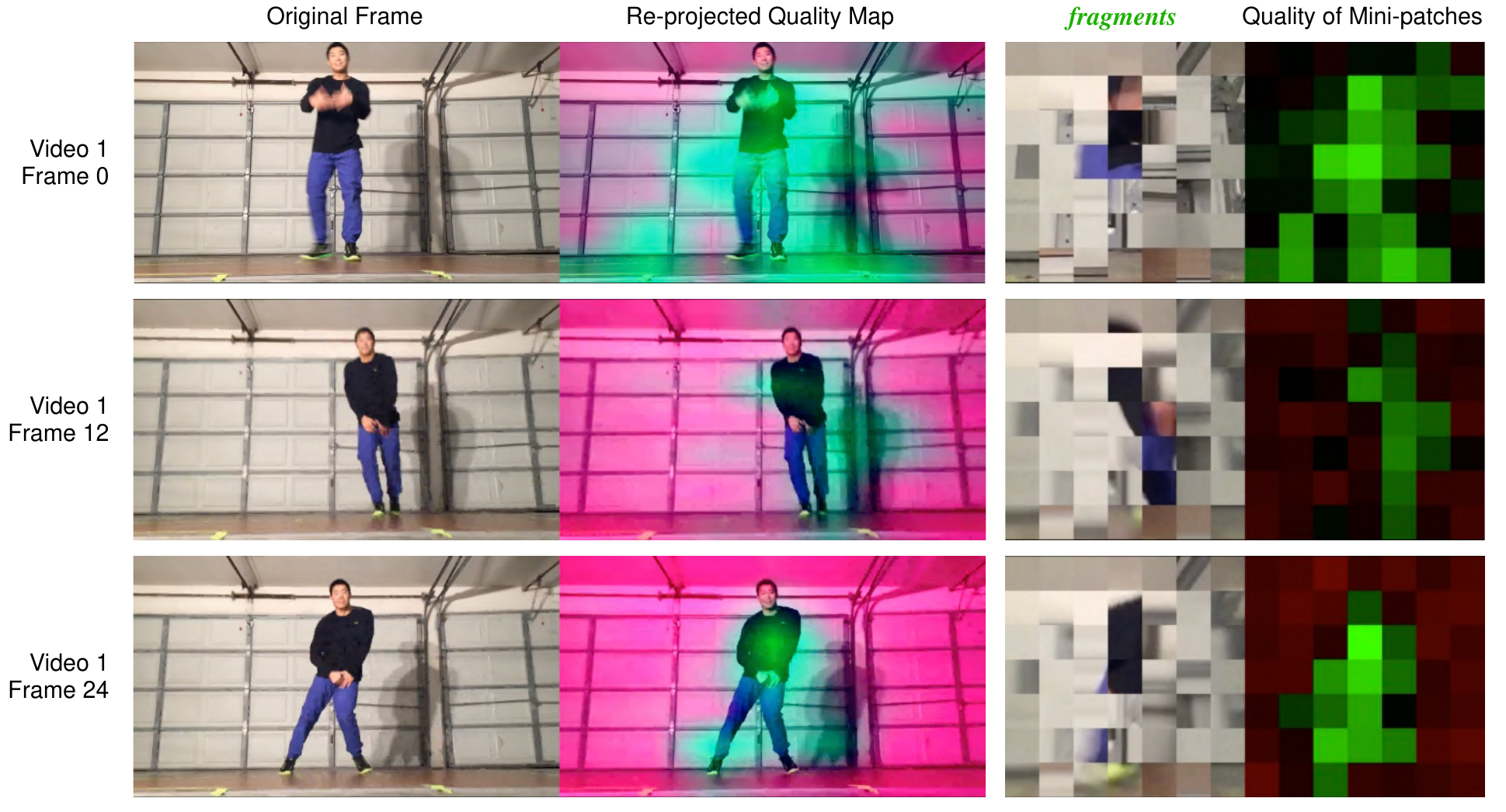}
    \vspace{-12pt}
    \caption{Spatial-temporal patch-wise local quality maps, where \textbf{\red{red}} areas refer to low predicted quality and \textbf{\green{green}} areas refer to high predicted quality. This sample video is a 1080P video selected from LIVE-VQC~\cite{vqc} dataset. Zoom in for clearer view.}
    \label{fig:6}
    \vspace{-16pt}
\end{figure}

\subsection{Qualitative Results: Local Quality Maps}

 The proposed IP-NLR head with patch-wise independent quality regression enables FAST-VQA to generate patch-wise local quality maps, which helps us to qualitatively evaluate what quality information can be learned in FAST-VQA. We show the patch-wise local quality maps and the re-projected frame quality maps for a 1080P video (from LIVE-VQC~\cite{vqc} dataset) in Fig.~\ref{fig:6}. As the patch-wise quality maps and re-projected quality maps in Fig.~\ref{fig:6} (column 2\&4) shows, FAST-VQA is sensitive to textural quality information and distinguishes between clear (Frame 0) and blurry textures (Frame 12/24). It demonstrates that FAST-VQA with \frag~(column 3) as input is sensitive to local texture quality. Furthermore, the qualities of the action-related areas are notably different from the background areas, showing that FAST-VQA effectively learns the global scene information and contextual relations in the video.

\section{Conclusions}

Our paper has shown that proposed \frag~are effective samples for video quality assessment (VQA) that better retain quality information in videos than naive sampling approaches, to tackle the difficulties as results of high computing and memory requirements when high-resolution videos are to be evaluated. Based on \frag, the proposed end-to-end FAST-VQA achieves higher efficiency ($-99.5\%$ FLOPs) and accuracy ($+10\%$ PLCC) simultaneously than existing state-of-the-art method PVQ on 1080P videos. We hope that the FAST-VQA can bring deep VQA methods into practical use for videos in any resolutions.

\section{Acknowledgement}
This study is supported under the RIE2020 Industry Alignment Fund – Industry Collaboration Projects (IAF-ICP) Funding Initiative, as well as cash and in-kind contribution from the industry partner(s).
\clearpage
%
%
\bibliographystyle{splncs04}
\bibliography{egbib}

\begin{thebibliography}{10}
\providecommand{\url}[1]{\texttt{#1}}
\providecommand{\urlprefix}{URL }
\providecommand{\doi}[1]{https://doi.org/#1}

\bibitem{vivit}
Arnab, A., Dehghani, M., Heigold, G., Sun, C., Lucic, M., Schmid, C.: Vivit: A
  video vision transformer. In: Proceedings of the IEEE/CVF International
  Conference on Computer Vision (ICCV). pp. 6836--6846 (October 2021)

\bibitem{activitynet}
Caba~Heilbron, F., Escorcia, V., Ghanem, B., Carlos~Niebles, J.: Activitynet: A
  large-scale video benchmark for human activity understanding. In: Proceedings
  of the IEEE/CVF Conference on Computer Vision and Pattern Recognition (CVPR)
  (June 2015)

\bibitem{gstvqa}
Chen, B., Zhu, L., Li, G., Lu, F., Fan, H., Wang, S.: Learning generalized
  spatial-temporal deep feature representation for no-reference video quality
  assessment. IEEE Transactions on Circuits and Systems for Video Technology
  (2021)

\bibitem{gru}
Cho, K., van Merrienboer, B., G{\"{u}}l{\c{c}}ehre, {\c{C}}., Bahdanau, D.,
  Bougares, F., Schwenk, H., Bengio, Y.: Learning phrase representations using
  {RNN} encoder-decoder for statistical machine translation. In: Proceedings of
  the 2014 Conference on Empirical Methods in Natural Language Processing
  ({EMNLP}). pp. 1724--1734. {ACL} (2014)

\bibitem{imagenet}
Deng, J., Dong, W., Socher, R., Li, L.J., Li, K., Fei-Fei, L.: Imagenet: A
  large-scale hierarchical image database. In: Proceedings of the IEEE/CVF
  Conference on Computer Vision and Pattern Recognition (CVPR). pp. 248--255
  (2009)

\bibitem{mvit}
Fan, H., Xiong, B., Mangalam, K., Li, Y., Yan, Z., Malik, J., Feichtenhofer,
  C.: Multiscale vision transformers. In: Proceedings of the IEEE/CVF
  International Conference on Computer Vision (ICCV). pp. 6824--6835 (October
  2021)

\bibitem{qualcomm}
Ghadiyaram, D., Pan, J., Bovik, A.C., Moorthy, A.K., Panda, P., Yang, K.C.:
  In-capture mobile video distortions: A study of subjective behavior and
  objective algorithms. IEEE Transactions on Circuits and Systems for Video
  Technology  \textbf{28}(9),  2061--2077 (2018)

\bibitem{mlsp}
G\"otz-Hahn, F., Hosu, V., Lin, H., Saupe, D.: Konvid-150k: A dataset for
  no-reference video quality assessment of videos in-the-wild. In: IEEE Access
  9. pp. 72139--72160. IEEE (2021)

\bibitem{ava}
Gu, C., Sun, C., Ross, D.A., Vondrick, C., Pantofaru, C., Li, Y.,
  Vijayanarasimhan, S., Toderici, G., Ricco, S., Sukthankar, R., Schmid, C.,
  Malik, J.: Ava: A video dataset of spatio-temporally localized atomic visual
  actions. In: Proceedings of the IEEE/CVF Conference on Computer Vision and
  Pattern Recognition (CVPR) (June 2018)

\bibitem{r3d}
Hara, K., Kataoka, H., Satoh, Y.: Learning spatio-temporal features with 3d
  residual networks for action recognition. In: Proceedings of the IEEE/CVF
  International Conference on Computer Vision (ICCV) Workshops. pp. 3154--3160
  (2017)

\bibitem{he2016residual}
He, K., Zhang, X., Ren, S., Sun, J.: Deep residual learning for image
  recognition. In: Proceedings of the IEEE/CVF Conference on Computer Vision
  and Pattern Recognition (CVPR). pp. 770--778 (2016)

\bibitem{kv1k}
Hosu, V., Hahn, F., Jenadeleh, M., Lin, H., Men, H., Szirányi, T., Li, S.,
  Saupe, D.: The konstanz natural video database (konvid-1k). In: Ninth
  International Conference on Quality of Multimedia Experience (QoMEX).
  pp.~1--6 (2017)

\bibitem{koniq}
Hosu, V., Lin, H., Sziranyi, T., Saupe, D.: Koniq-10k: An ecologically valid
  database for deep learning of blind image quality assessment. IEEE
  Transactions on Image Processing  \textbf{29},  4041--4056 (2020)

\bibitem{crop1}
Kang, L., Ye, P., Li, Y., Doermann, D.: Convolutional neural networks for
  no-reference image quality assessment. Proceedings of the IEEE/CVF Conference
  on Computer Vision and Pattern Recognition (CVPR)  (2014)

\bibitem{crop2}
Kang, L., Ye, P., Li, Y., Doermann, D.: Simultaneous estimation of image
  quality and distortion via multi-task convolutional neural networks. IEEE
  international conference on image processing (ICIP)  (2015)

\bibitem{k400data}
Kay, W., Carreira, J., Simonyan, K., Zhang, B., Hillier, C., Vijayanarasimhan,
  S., Viola, F., Green, T., Back, T., Natsev, A., Suleyman, M., Zisserman, A.:
  The kinetics human action video dataset. ArXiv  \textbf{abs/1705.06950}
  (2017)

\bibitem{musiq}
Ke, J., Wang, Q., Wang, Y., Milanfar, P., Yang, F.: Musiq: Multi-scale image
  quality transformer. In: Proceedings of the IEEE/CVF International Conference
  on Computer Vision (ICCV). pp. 5148--5157 (October 2021)

\bibitem{deepvqa}
Kim, W., Kim, J., Ahn, S., Kim, J., Lee, S.: Deep video quality assessor: From
  spatio-temporal visual sensitivity to a convolutional neural aggregation
  network. In: Proceedings of the European Conference on Computer Vision (ECCV)
  (2018)

\bibitem{vit}
Kolesnikov, A., Dosovitskiy, A., Weissenborn, D., Heigold, G., Uszkoreit, J.,
  Beyer, L., Minderer, M., Dehghani, M., Houlsby, N., Gelly, S., Unterthiner,
  T., Zhai, X.: An image is worth 16x16 words: Transformers for image
  recognition at scale (2021)

\bibitem{tlvqm}
Korhonen, J.: Two-level approach for no-reference consumer video quality
  assessment. IEEE Transactions on Image Processing  \textbf{28}(12),
  5923--5938 (2019)

\bibitem{cnntlvqm}
Korhonen, J., Su, Y., You, J.: Blind natural video quality prediction via
  statistical temporal features and deep spatial features. In: Proceedings of
  the 28th ACM International Conference on Multimedia. p. 3311–3319. MM '20,
  Association for Computing Machinery, New York, NY, USA (2020)

\bibitem{vsfa}
Li, D., Jiang, T., Jiang, M.: Quality assessment of in-the-wild videos. In:
  Proceedings of the 27th ACM International Conference on Multimedia. p.
  2351–2359. MM '19, Association for Computing Machinery, New York, NY, USA
  (2019)

\bibitem{mdtvsfa}
Li, D., Jiang, T., Jiang, M.: Unified quality assessment of in-the-wild videos
  with mixed datasets training. International Journal of Computer Vision
  \textbf{129}(4),  1238--1257 (2021)

\bibitem{sfa}
Li, D., Jiang, T., Lin, W., Jiang, M.: Which has better visual quality: The
  clear blue sky or a blurry animal? IEEE Transactions on Multimedia
  \textbf{21}(5),  1221--1234 (2019)

\bibitem{tpqi}
Liao, L., Xu, K., Wu, H., Chen, C., Sun, W., Yan, Q., Lin, W.: Exploring the
  effectiveness of video perceptual representation in blind video quality
  assessment. In: Proceedings of the 30th ACM International Conference on
  Multimedia (ACM MM) (2022)

\bibitem{swin2d}
Liu, Z., Lin, Y., Cao, Y., Hu, H., Wei, Y., Zhang, Z., Lin, S., Guo, B.: Swin
  transformer: Hierarchical vision transformer using shifted windows. arXiv
  preprint arXiv:2103.14030  (2021)

\bibitem{swin3d}
Liu, Z., Ning, J., Cao, Y., Wei, Y., Zhang, Z., Lin, S., Hu, H.: Video swin
  transformer. arXiv preprint arXiv:2106.13230  (2021)

\bibitem{brisque}
Mittal, A., Moorthy, A.K., Bovik, A.C.: No-reference image quality assessment
  in the spatial domain. IEEE Transactions on Image Processing
  \textbf{21}(12),  4695--4708 (2012)

\bibitem{viideo}
Mittal, A., Saad, M.A., Bovik, A.C.: A completely blind video integrity oracle.
  IEEE Transactions on Image Processing  \textbf{25}(1),  289--300 (2016)

\bibitem{cvd}
Nuutinen, M., Virtanen, T., Vaahteranoksa, M., Vuori, T., Oittinen, P.,
  Häkkinen, J.: Cvd2014—a database for evaluating no-reference video quality
  assessment algorithms. IEEE Transactions on Image Processing  \textbf{25}(7),
   3073--3086 (2016)

\bibitem{vbliinds}
Saad, M.A., Bovik, A.C., Charrier, C.: Blind image quality assessment: A
  natural scene statistics approach in the dct domain. IEEE Transactions on
  Image Processing  \textbf{21}(8),  3339--3352 (2012)

\bibitem{vqc}
Sinno, Z., Bovik, A.C.: Large-scale study of perceptual video quality. IEEE
  Transactions on Image Processing  \textbf{28}(2),  612--627 (2019)

\bibitem{irnv2}
Szegedy, C., Ioffe, S., Vanhoucke, V., Alemi, A.A.: Inception-v4,
  inception-resnet and the impact of residual connections on learning. In:
  Proceedings of the Thirty-First AAAI Conference on Artificial Intelligence.
  p. 4278–4284. AAAI'17, AAAI Press (2017)

\bibitem{deit}
Touvron, H., Cord, M., Douze, M., Massa, F., Sablayrolles, A., J'egou, H.:
  Training data-efficient image transformers \& distillation through attention.
  In: Proceedings of the International Conference on Machine Learning (ICML)
  (2021)

\bibitem{rapique}
Tu, Z., Chen, C.J., Wang, Y., Birkbeck, N., Adsumilli, B., Bovik, A.C.:
  Efficient user-generated video quality prediction. In: 2021 Picture Coding
  Symposium (PCS). pp.~1--5 (2021)

\bibitem{videval}
Tu, Z., Wang, Y., Birkbeck, N., Adsumilli, B., Bovik, A.C.: Ugc-vqa:
  Benchmarking blind video quality assessment for user generated content. IEEE
  Transactions on Image Processing  \textbf{30},  4449--4464 (2021)

\bibitem{rfugc}
Wang, Y., Ke, J., Talebi, H., Yim, J.G., Birkbeck, N., Adsumilli, B., Milanfar,
  P., Yang, F.: Rich features for perceptual quality assessment of ugc videos.
  In: Proceedings of the IEEE/CVF Conference on Computer Vision and Pattern
  Recognition (CVPR). pp. 13435--13444 (June 2021)

\bibitem{ytugc}
Yim, J.G., Wang, Y., Birkbeck, N., Adsumilli, B.: Subjective quality assessment
  for youtube ugc dataset. In: 2020 IEEE International Conference on Image
  Processing (ICIP). pp. 131--135 (2020)

\bibitem{paq2piq}
Ying, Z.a., Niu, H., Gupta, P., Mahajan, D., Ghadiyaram, D., Bovik, A.: From
  patches to pictures (paq-2-piq): Mapping the perceptual space of picture
  quality. arXiv preprint arXiv:1912.10088  (2019)

\bibitem{pvq}
Ying, Z., Mandal, M., Ghadiyaram, D., Bovik, A.: Patch-vq: 'patching up' the
  video quality problem. In: Proceedings of the IEEE/CVF Conference on Computer
  Vision and Pattern Recognition (CVPR)). pp. 14019--14029 (June 2021)

\bibitem{lsctphiq}
You, J.: Long short-term convolutional transformer for no-reference video
  quality assessment. In: Proceedings of the 29th ACM International Conference
  on Multimedia. p. 2112–2120. MM '21, Association for Computing Machinery,
  New York, NY, USA (2021)

\bibitem{cnn+lstm}
You, J., Korhonen, J.: Deep neural networks for no-reference video quality
  assessment. In: Proceedings of the IEEE International Conference on Image
  Processing (ICIP). pp. 2349--2353 (2019)

\bibitem{dbcnn}
Zhang, W., Ma, K., Yan, J., Deng, D., Wang, Z.: Blind image quality assessment
  using a deep bilinear convolutional neural network. IEEE Transactions on
  Circuits and Systems for Video Technology  \textbf{30}(1),  36--47 (2020)

\end{thebibliography}
\end{document}